\documentclass[conference]{IEEEtran}
\usepackage{times}

\usepackage[numbers]{natbib}
\usepackage{graphics}
\usepackage[utf8]{inputenc}
\usepackage{booktabs}
\usepackage{amsmath} 
\usepackage{amssymb}  
\usepackage{amsfonts}
\usepackage{mathtools}
\usepackage[font=small]{caption}
\usepackage{booktabs}
\usepackage{wrapfig}
\usepackage{arrayjob}
\usepackage{flushend}
\usepackage{pifont}
\usepackage{bbm}
\usepackage{transparent}
\usepackage{xcolor}

\def\BibTeX{{\rm B\kern-.05em{\sc i\kern-.025em b}\kern-.08em
    T\kern-.1667em\lower.7ex\hbox{E}\kern-.125emX}}

\usepackage{siunitx}
\usepackage{bm}

\newcommand{\R}{\mathbb{R}}
\DeclareMathAlphabet\mathbfcal{OMS}{cmsy}{b}{n}

\DeclarePairedDelimiterX{\infdivx}[2]{(}{)}{%
  #1\;\delimsize|\delimsize|\;#2%
}

\usepackage{caption}
\usepackage{booktabs}

\usepackage{caption}
\usepackage{subcaption}
\usepackage{booktabs}
\usepackage{xcolor}
\usepackage{colortbl}
\usepackage{pgfplots}
\pgfplotsset{compat=1.16}
\usetikzlibrary{intersections}
\usepgfplotslibrary{fillbetween}
\definecolor{lightblue}{RGB}{173,216,230} 
\definecolor{lightred}{RGB}{255,182,193}
\usepackage{grffile}
\usepackage[hidelinks]{hyperref}
\usepackage[capitalise]{cleveref}
\usepackage{graphicx}
\usepackage{svg}

\usepackage{mathtools}

\begin{document}

\title{Self-supervised Multi-future Occupancy Forecasting \\ for Autonomous Driving}





\author{\authorblockN{Bernard Lange\authorrefmark{1},
Masha Itkina\authorrefmark{1}, 
Jiachen Li\authorrefmark{2} and
Mykel J. Kochenderfer\authorrefmark{1}}
\authorblockA{ 
\authorrefmark{1}Stanford University, \authorrefmark{2}University of
California, Riverside}
}

\maketitle

\begin{abstract}
Environment prediction frameworks are critical for the safe navigation of autonomous vehicles (AVs) in dynamic settings. LiDAR-generated occupancy grid maps (L-OGMs) offer a robust bird's-eye view for the scene representation, enabling self-supervised joint scene predictions while exhibiting resilience to partial observability and perception detection failures. Prior approaches have focused on deterministic L-OGM prediction architectures within the grid cell space. While these methods have seen some success, they frequently produce unrealistic predictions and fail to capture the stochastic nature of the environment. Additionally, they do not effectively integrate additional sensor modalities present in AVs. Our proposed framework, Latent Occupancy Prediction (LOPR), performs stochastic L-OGM prediction in the latent space of a generative architecture and allows for conditioning on RGB cameras, maps, and planned trajectories. We decode predictions using either a single-step decoder, which provides high-quality predictions in real-time, or a diffusion-based batch decoder, which can further refine the decoded frames to address temporal consistency issues and reduce compression losses. Our experiments on the nuScenes and Waymo Open datasets show that all variants of our approach qualitatively and quantitatively outperform prior approaches.
\end{abstract} 

\IEEEpeerreviewmaketitle

\section{Introduction}
\label{sec:intro}
Accurate environment prediction algorithms are essential for autonomous vehicle (AV) navigation in urban settings.
Experienced drivers understand scene semantics and recognize the intent of other agents to anticipate their trajectories and safely navigate to their destination. To replicate this process in AVs, many environment prediction approaches have been proposed, employing different environment representations and modeling assumptions~\citep{itkina2019dynamic, lange2020attention, toyungyernsub2021double, chai2019multipath, choi2021shared, mahjourian2022occupancy,lange2023scene}. 

The modern AV stack comprises a mixture of expert-designed and learned modules, such as 3D object detection, tracking, motion forecasting, and planning, each developed independently. In the case of learned systems, development involves using curated labels provided by human annotators and other perception systems.
For environmental reasoning, object-based prediction algorithms are often used, which rely on the perception system to create a vectorized representation of the scene with defined agents and environmental features~\citep{chai2019multipath, nayakanti2022wayformer}. However, this approach has multiple limitations.
First, it often generates marginalized future trajectories for each individual agent, rather than a holistic scene prediction including agent interactions, which complicates integration with planning modules~\cite{chen2022scept}.
Second, this approach does not take sensor measurements into account and depends solely on object detection algorithms that may fail in suboptimal conditions~\citep{delecki2022we, dreissig2023survey}. 
Third, reliance on labeled data, sourced from both human annotators and perception systems, constrains the dataset size and incurs higher costs. 
These drawbacks render the AV stack susceptible to cascading failures and can lead to poor generalization in unforeseen scenarios. Such limitations underscore the need for complementary environment modeling approaches that do not rely on error-prone and expensive labeling schemes.


In response to these challenges, occupancy grid maps generated from LiDAR measurements (L-OGMs) have gained popularity as a form of scene representation for prediction. The popularity is due to their minimal data preprocessing requirements, eliminating the need for manual labeling, the ability to model the joint prediction of the scene with an arbitrary number of agents, and robustness to partial observability and detection failures~\cite{itkina2019dynamic, lange2020attention}.
We focus on ego-centric L-OGM prediction generated using uncertainty-aware occupancy state estimation approaches~\citep{elfes1989using}. Due to its generality and ability to scale with unlabeled data, we hypothesize that L-OGM prediction alongside RGB video prediction could also serve as an unsupervised pre-training objective, i.e., a foundational model, for autonomous driving.

\begin{figure}[t]
\centering
\includegraphics[width=\columnwidth]{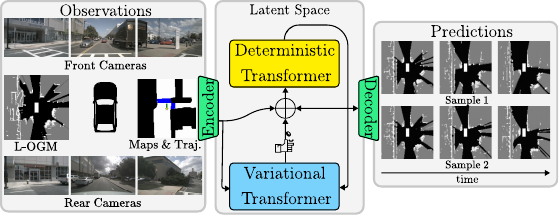} 
\label{fig:figure-label}

\caption{Latent Occupancy PRediction (LOPR) is a self-supervised stochastic prediction framework that forecasts occupancy grid maps within the latent space of a generative model. It consists of deterministic and variational transformer modules conditioned on occupancy grids, images, maps, and the planned trajectory. LOPR forecasts multiple plausible futures for the entire scene.} 
\label{fig:overview}
\end{figure}
The task of L-OGM prediction is typically framed as self-supervised sequence-to-sequence learning. ConvLSTM-based architectures have been used predominantly in previous work for this task due to their ability to handle spatiotemporal sequences~\citep{itkina2019dynamic, schreiber2019long, lange2020attention, toyungyernsub2021double}. These approaches are optimized end-to-end in grid cell space, do not account for the stochasticity present in the scene, and neglect other available modalities, e.g., RGB cameras around the vehicle, maps, and the planned trajectory. As a result, they often suffer from unrealistic and blurry predictions. 

In this work, we address the limitations of previous approaches by proposing a stochastic L-OGM prediction framework that operates within the latent space of a generative model~\citep{larsen2016autoencoding}. Generative models are known for providing a compressed representation, while producing high-quality samples~\citep{goodfellow2014generative, kingma2013auto}. With the use of generative models, we can minimize redundancies in the representation, allowing the prediction network to focus computation on the most critical aspects of the task~\citep{yan2021videogpt}. Vector-quantized variational models~\citep{van2017neural} combined with autoregressive transformers~\citep{vaswani2017attention} have demonstrated significant success. However, it often comes at the expense of increased inference time, driven by the large number of discrete tokens needed to effectively capture the task~\citep{rakhimov2020latent}. In this work, we leverage lower dimensional continuous representations to enable real-time performance.

Within the latent space trained on L-OGMs, our framework employs an autoregressive transformer-based architecture consisting of two modules, called sequentially at each time step: a variational module that models the stochasticity of the scene and a deterministic module that predicts the next time step. Both are conditioned on past L-OGM encodings and other modalities if available, such as camera images, maps, and the planned trajectory, as shown in \cref{fig:overview}. Predictions are decoded one by one using a single-step decoder, which provides high-quality predictions in real-time that optionally can be refined with a diffusion-based batch decoder. The diffusion-based batch decoder addresses the temporal consistency issues associated with single-step decoders~\cite{hu2023gaia} and mitigates compression losses by conditioning on prior rasterized L-OGMs, at the cost of real-time feasibility.

Experiments on nuScenes~\citep{caesar2020nuscenes} and the Waymo Open Dataset~\citep{sun2020scalability} show quantitative and qualitative improvements over prior approaches, including recent state-of-the-art discrete transformer-based approaches. Our framework forecasts diverse futures and infers unobserved agents. It also leverages other sensor modalities for more accurate predictions, such as observing oncoming vehicles in a camera feed beyond the visible region of the L-OGMs. Our contributions are:
\begin{itemize} 
\item We introduce a framework called \textbf{L}atent \textbf{O}ccupancy \textbf{PR}ediction (LOPR) for stochastic L-OGM prediction in the latent space of a generative model conditioned on other sensor modalities, such as RGB cameras, maps, and the planned AV trajectory.
\item We propose a variational-based transformer model that captures the stochasticity of the surrounding scene while remaining real-time feasible.
\item We define a diffusion-based batch decoder that refines single-frame decoder outputs to address temporal consistency issues and reduce compression losses. 
\item Through experiments on the nuScenes~\citep{caesar2020nuscenes} and Waymo Open Dataset~\citep{sun2020scalability}, we demonstrate that LOPR surpasses prior L-OGM prediction methods and highlight the positive impact of incorporating additional input modalities.

\end{itemize}
\vspace{-0.25cm}


\section{Related Work}
\label{sec:related}
\textbf{OGM Prediction.}
The majority of prior work in OGM prediction generates OGMs with LiDAR measurements (L-OGMs) and uses an adaptation on the recurrent neural network (RNN) with convolutions \citep{toyungyernsub2022dynamics,Toyungyernsub2024predicting}. \citet{dequaire2018deep} tracked objects through occlusions and predicted future binary OGMs with an RNN and a spatial transformer. \citet{schreiber2019long} provided dynamic occupancy grid maps (DOGMas) with cell-wise velocity estimates as input to a ConvLSTM for environment prediction from a stationary platform. \citet{schreiber2021dynamic} then extended this work to forecast DOGMas in a moving ego-vehicle setting. \citet{mohajerin2019multi} applied a difference learning approach to predict OGMs as seen from the coordinate frame of the first observed time step. \citet{itkina2019dynamic} used the PredNet ConvLSTM architecture~\cite{lotter2016deep} to achieve ego-centric OGM prediction. \citet{lange2020attention} reduced the blurring and the gradual disappearance of dynamic obstacles in the predicted grids by developing an attention augmented ConvLSTM mechanism. Concurrently, \citet{toyungyernsub2021double} addressed obstacle disappearance with a double-prong framework assuming knowledge of the static and dynamic obstacles.
An alternative approach predicts occupancy grid maps from vectorized object data~\cite{mahjourian2022occupancy} or a mixture of vectorized object data and sensor measurements~\cite{zheng2023occworld}. Similar to representations in common trajectory prediction techniques, these methods require substantial labeling efforts~\cite{ettinger2021large, qi2021offboard, tian2023occ3d}. Unlike prior work, we perform self-supervised multi-future L-OGM predictions in the latent space of generative models conditioned on additional sensor modalities without the need for manual labeling. 


\textbf{Representation Learning in Robotics and Autonomous Driving}. The objective of representation learning is to identify low-dimensional representations that make it easier to achieve the desired performance on a task. Many robotics applications use architectures such as the autoencoder~(AE)~\cite{hinton1993autoencoders}, the variational autoencoder~(VAE)~\cite{kingma2013auto}, the generative adversarial network~(GAN)~\cite{goodfellow2014generative},
and the vector quantized variational autoencoder (VQ-VAE)~\cite{van2017neural}. Latent spaces have been used to learn latent dynamics from pixels~\citep{ha2018recurrent}, output video predictions~\citep{babaeizadeh2017stochastic}, generate trajectories~\citep{hafner2019dream}, and learn autonomous driving neural simulators~\citep{kim2021drivegan, hu2023gaia}. Large-scale video prediction architectures have used discrete representations provided by the VQ-VAE~\cite{van2017neural} with a causal transformer~\cite{vaswani2017attention, rakhimov2020latent, hu2023gaia}. However, these models remain prohibitively expensive to train and sample from due to large number of discrete tokens required. We present a method that performs multi-future L-OGM prediction entirely in the continuous latent space of VAE-GAN generative model in real time. 
\begin{figure*}[t]
    \centering
    \includegraphics[width=\textwidth]{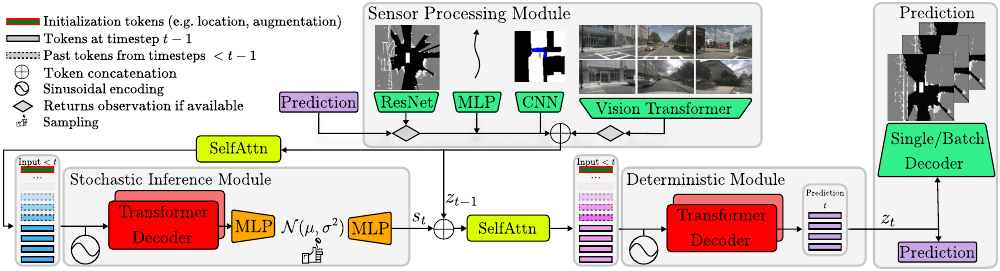} 
  \caption{The LOPR framework, comprising sensor processing, stochastic inference, and prediction modules. The sensor processing module encodes all sensor modalities. The L-OGM and RGB camera encoders are pretrained as described in \cref{sec:unsupervised} and in \cref{sec:modality}. The inference module captures the scene's stochasticity (\cref{sec:prediction}). In the prediction module, we forecast the next time step's L-OGM embedding. At each time step, the most recent predictions are autoregressively provided to the inference and prediction modules.}   
  \label{fig:LOPR_details}
\end{figure*}

\section{LOPR: Latent Occupancy PRediction}
We propose the \textbf{Latent Occupancy Prediction (LOPR)} framework, designed to generate stochastic scene predictions represented as ego-centric L-OGMs. A 2D L-OGM grid  \(x \in \mathbb{R}^{H \times W}\)  represents a bird’s-eye-view map with dimensions (H, W), where each cell encodes the probability of occupancy. The grid is generated using projected LiDAR sensor measurements with segmented ground. The task of stochastic prediction involves learning a distribution over future occupancy grids conditioned on the observed grids, \( p(x_{>T_O} \mid x_{\leq T_O}) \), where \( T_O \) denotes the observation horizon.

LOPR separates the task into (1) learning an L-OGM representation and (2) making predictions in the latent space of a generative model. In the representation learning phase, a VAE-GAN is trained to learn an L-OGM latent space. During the prediction stage, our framework uses an autoregressive transformer-based architecture, comprising both deterministic and variational decoder models. At each time step, a sample is drawn from the variational transformer and then passed to the deterministic transformer to forecast the next L-OGM embedding. Predictions are conditioned on past L-OGMs encodings and other available modalities, such as camera images, maps, and the planned trajectory. The encoders for maps and planned trajectories are trained alongside the prediction framework, while for the image encoder, we use a pre-trained DINOv2-based model~\citep{oquab2023dinov2}. Predictions are decoded using a single-step decoder, which provides high-quality predictions in real-time that optionally can be refined with a diffusion-based batch decoder. \cref{fig:LOPR_details} summarizes the framework.

\subsection{Representation Learning}
\label{sec:unsupervised}
The latent space for L-OGMs is obtained by training an encoder and a decoder. Given an L-OGM grid $x$, the encoder outputs a low-dimensional latent representation $z \in\R^{c \times h \times w}$ with dimensions $(h, w)$ and depth $c$. This representation is reconstructed by a decoder to $\hat{x}\in\R^{H \times W}$. The framework integrates concepts from $\beta$-VAE and GAN~\citep{larsen2016autoencoding, kim2021drivegan}. In $\beta$-VAE, the objective consists of the reconstruction and regularization losses:
\begin{align}
\label{eqn:vae}
\mathcal{L}_{\text{VAE}}=-\mathbb{E}_{q\left(z \mid x\right)}\left[\log p\left(x \mid z\right)\right] +\beta D_{\text{KL}}\left(q\left(z \mid x\right) ||\; p\left(z\right)\right),
\end{align}
where $q\left(z \mid x\right)$ and $p\left(x \mid z\right)$ are the outputs of the encoder and decoder respectively, $p\left(z\right)$ is the unit Gaussian prior, and $D_{\text{KL}}$ represents the Kullback-Leibler divergence. The reconstruction loss is an average of the perceptual loss~\citep{zhang2018unreasonable} and the mean squared error.
In the GAN step, the same decoder serves as the generator, and a discriminator classifies whether samples originate from the training set. This framework uses minimax optimization for the following objective~\cite{goodfellow2014generative}:
\begin{align}
\label{eqn:gan}
\mathcal{L}_{\text{GAN}}=\mathbb{E}_{x \sim p_{\text{data}}}\left[\log \mathcal{D}(x)\right] + \mathbb{E}_{\hat{x} \sim p_{\text{model}}}\left[\log \left(1-\mathcal{D}(\hat{x})\right)\right],
\end{align}
where $\mathcal{D}$ is the discriminator. The final loss is $\mathcal{L}_{\text{rep}}=\mathcal{L}_{\text{VAE}}+\mathcal{L}_{\text{GAN}}$ and follows the implementation described by~\citet{karras2020analyzing} and~\citet{kim2021drivegan}.

\subsection{Stochastic L-OGM Sequence Prediction}
\label{sec:prediction}
Given the pre-trained L-OGM latent space, we train a stochastic sequence prediction network that receives a history of observations and outputs a distribution over a potential future embedding $p_{\theta}(z_{t} \mid z_{<t})$, where $z_{<t}$ represents the compressed L-OGM representations over the last $t$ time steps, and $\theta$ are the network weights. 
The environment prediction task is inherently multimodal, and the latent vectors contributing to this stochasticity are unobservable. We introduce a variable $s_t$ to capture this stochasticity at timestep $t$ of the sequence and extend our model to $p_{\theta}(z_{t} \mid z_{<t}, s_{\leq t})$~\citep{babaeizadeh2017stochastic}. During training, we extract the true posterior using an inference network $s_t \sim q_{\phi}(s_t \mid z_{\leq t})$ which has access to the $z_t$, representation of L-OGM at timestep $t$. While at test time, we sample from a learned prior $s_t \sim p_{\gamma}(s_t \mid z_{<t})$ as we are attemping to predict $z_t$. This process is autoregressively repeated for $T_F$ future steps assuming access for $T_O$ past observations. The framework is optimized using the variational lower bound objective~\cite{kingma2013auto}: 
\begin{align}
\label{eqn:elbo}
\mathcal{L}=&-\sum_{t=T_O}^{T_O + T_F} \Bigg(\mathbb{E}_{q_\phi\left(s_{\leq t} \mid z_{\leq t}\right)}\left[\log p_\theta\left(z_{t} \mid z_{<t}, s_{\leq t}\right)\right] \\ \nonumber &-D_{K L}\left(q_\phi\left(s_{t} \mid z_{\leq t}\right) || \; p_\gamma\left(s_{t} \mid z_{<t}\right)\right)\Bigg),
\end{align}


where the prediction network and prior autoregressively receive previous predicted embeddings, whereas the posterior takes in only the ground truth.

LOPR is implemented using a transformer decoder-based architecture, comprising a deterministic module $\mathcal{P}_{\theta}$ and two inference networks $\mathcal{Q}_{\phi}$ and $\mathcal{Q}_{\gamma}$ for the prior and posterior, respectively. The positional information is provided through a sinusoidal positional encoding~\citep{vaswani2017attention}.
At each time step $t$, $s_t$ is sampled from the inference network $\mathcal{Q}$, which outputs the parameters of the Gaussian distribution:
\begin{align}
    \label{eqn:inference}
    \mu, \; \sigma &= \mathcal{Q}(z_{\leq \; \text{or} \; < t}) \\ 
    s_{t} &\sim \mathcal{N}\left(\mu, \sigma^2\right), 
\end{align}
where $s_t$ is drawn from $\mathcal{Q}_{\gamma}$ at test time and from $\mathcal{Q}_{\phi}$ at training time, as explained above. Then $z_{t-1}$ is attended with $s_t$ and provided to the deterministic decoder where all past representations are incorporated: %
\begin{align}
z_{t} = \mathcal{P}_{\theta}(\text{SelfAttn}(z_{< t}, s_{\leq t})).    
\end{align}
In the above operations, each $z_t \in \R^{c \times h \times w}$ is split along its spatial dimensions into $k$ patches, which are then flattened~\cite{dosovitskiy2020image}. Each token has dimension $\frac{chw}{k}$, thereby also facilitating spatial attention and optimizing the parameter count in the attention layer. This operation is applied to both the deterministic decoder and the inference networks. In the final step, the predicted compressed representations are concatenated, reshaped back to their original dimensions, and then provided to the decoder from \cref{sec:unsupervised}. 

\subsection{Diffusion-based Batch Decoder}
\label{sec: 3d_decoder}

We can decode each $z_t$ independently using the single-frame decoder outlined in \cref{sec:unsupervised} to obtain high-quality predictions in real time. However, this approach can lead to poor temporal consistency and compression losses~\cite{hu2023gaia}. They manifest as unrealistic changes in the distribution of occupied cells over time and poor pixel-wise accuracy, particularly in the first predicted frames that should retain most of the static details from the observations.

We address these issues by refining $\hat{x}_{t-\Delta:t}$ from the single-frame decoder in batches with a diffusion-based batch decoder, where $\Delta$ is the number of frames. $\Delta$ is fixed and is not a function of batch size used at inference. At train time, the batch decoder is conditioned on ground truth decoded frames $\hat{x}_{t-\Delta:t}$ and a original rasterized frame preceding the sequence $x_{t-\Delta-1}$.
We follow a standard video diffusion formulation that uses a 3D-UNet as a denoising model and minimizes the mean-squared error between the predicted and ground truth noises~\cite{ho2022video}. The model is trained to refine the decoded ground truth frames $\hat{x}_{t-\Delta:t}$ to more closely match $x_{t-\Delta:t}$.
At test time, decoded frames and the preceding rasterized frame are reconstructed from predicted embeddings, with the exception of the first prediction, where the preceding frame is a rasterized observation, allowing static details to be preserved throughout the predicted sequence.

\subsection{Conditioning on Other Sensor Modalities}
\label{sec:modality}
LOPR can be conditioned on maps, the planned trajectory, and observed RGB camera images. We assume access to maps for the entire planned trajectory. Each input modality is first embedded as described below and then integrated into the framework using the self-attention mechanism before being provided to deterministic and inference networks. 

\textbf{Maps and Planned AV Trajectory.}
The map $m\in\R^{3 \times W \times H}$ comprises the drivable area, stop lines, and pedestrian crossings within the ego frame, represented using a rasterized format. The planned trajectory $tr \in\R^{3 \times T}$ includes position $(x,y,z)$ for the entire sequence, normalized with respect to the ego position. Maps and the planned trajectory are processed with commonly used convolutional and fully connected networks. 


\textbf{RGB Camera.}
The camera observation, denoted as $c_t\in\R^{N \times 3 \times W \times H}$, encompasses views from $N$ RGB cameras surrounding the vehicle. They offer important semantic information not available in L-OGMs. They can: 1) distinguish whether occupied cells are dynamic agents (like cars), including their type and orientation, or static environmental elements, and 2) provide insights into the state of the environment beyond the area observed in the fixed-size L-OGM. 
\begin{figure}[t]
    \centering
    \includegraphics[width=8 cm]{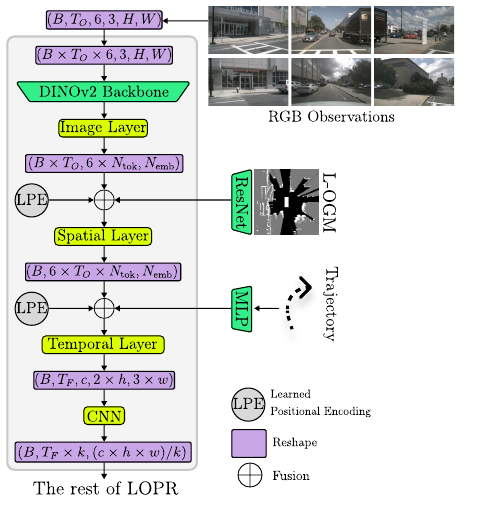} 
  \caption{Vision Transformer-based RGB Camera Encoder. RGB camera data is processed through the pre-trained DINOv2 backbone, subsequently passing through a series of attention layers. These layers aggregate information within each view (image layer), across different views (spatial layer), and throughout all observed timesteps (temporal layer). The spatial and temporal layers also include the learned positional embedding and are conditioned on the L-OGM embeddings and the planned trajectory, respectively.}   
  \label{fig:image_module}
  \vspace{-1.5em}
\end{figure}
Incorporating RGB inputs into the self-supervised perception task is challenging due to the limited size of the AV perception datasets. To address this, we use the pretrained DINOv2~\cite{oquab2023dinov2} backbone and finetune the module on a short deterministic L-OGM prediction task. The model is conditioned on observed L-OGMs and planned trajectory embeddings, which accelerates convergence. We also observe that adding trajectory embeddings significantly reduces uncertainty in future ego motion, enhancing the effectiveness of deterministic fine-tuning. This approach encourages the extraction of visual information useful for downstream stochastic prediction.
We note that DINOv2 is pre-trained on unlabeled data, aligning with the motivation of this work, unlike the commonly used ResNet-based backbones from object detection, which rely on manually labeled datasets.

\cref{fig:image_module} shows the image processing module $\mathcal{I}_{\beta}$. Each image is embedded using a DINOv2 backbone. It is followed by a series of attention modules: 1)~The image layer, which aggregates tokens from a single view. 2) The spatial layer, which collects embeddings from each perspective around the vehicle along with a corresponding L-OGM embedding. 3) The temporal layer, which aggregates information across all observed time steps and the planned AV trajectory. We found that adding planned AV trajectory is beneficial but not necessary. Finally, the tokens from each time step are concatenated along the spatial dimensions and then processed through a convolutional layer to produce $z_{\text{cam}}$. The output tokens are then segmented back into patches, flattened, and integrated into the framework using the self-attention mechanism.

\subsection{Conditioning on Other Information}
\label{sec:other}
\textbf{Locations.}
Considering the diversity of locations in nuScenes~\cite{caesar2020nuscenes}, we append a one-hot location encoding (Singapore or USA) to the start of the observations sequence. 

\textbf{Sequence Augmentation.}
The open-source perception datasets are relatively small compared to vectorized trajectory datasets. To increase the number of samples, we implement a series of augmentations (e.g. mirror reflections, rotations, and reversing the sequence). Recognizing that these augmentations might adversely affect the prediction correctness (e.g. potentially resulting in predictions that do not adhere to driving rules), we attach a one-hot encoding of the augmentation type at the beginning of the sequence.

\subsection{Prediction Extrapolation}
\label{sec:extrapolation}
We extend the prediction horizon at test time using a sliding-window approach, treating the last predicted frames as observations and repeating the process. However, unlike maps and planned trajectory modalities, we neither have access to nor predict future camera observations. Hence, we randomly drop out the image embeddings enabling robust extrapolation beyond the training-time prediction horizon. 
\section{Experiments}
\label{sec:experiments}
We evaluate our framework by analyzing the pre-trained latent space, evaluating its performance in environment prediction tasks, and examining the impact of additional sensor modalities on the predictions quality. 


\subsection{Dataset}
We use the nuScenes Dataset~\citep{caesar2020nuscenes} and the Waymo Open Dataset~\citep{sun2020scalability}.
nuScenes contains 5.5 hours of data collected in Boston and Singapore. Waymo Open Dataset~\citep{sun2020scalability} provides 6.4 hours of data compiled in San Francisco, Phoenix, and Mountain View. Both datasets include measurements from RGB cameras around the vehicle, LiDAR(s), and maps. 


\textbf{Data Representation}: We generate L-OGMs in the ego vehicle frame using a ground-segmented LiDAR point cloud. 
The OGM dimensions are $H \times W = 128 \times 128$ with a \SI[parse-numbers=false]{0.\overline{3}}{\metre} resolution, corresponding to a \SI{42.7}{\metre} $\times$ \SI{42.7}{\metre} grid. 
RGB images and maps are downscaled to $224 \times 224$ and $128 \times 128$ respectively. 
During sequence prediction training, we provide 5 past L-OGMs (\SI{0.5}{\second}) as observations alongside other sensor modalities, and forecast for 15 future frames (\SI{1.5}{\second}) at \SI{10}{\hertz}.
We also extend the prediction horizon to 30 frames (\SI{3.0}{\second}) to evaluate the extrapolation capabilities of our framework.

\subsection{Architecture and Training Details}
\label{sec:training}
\textbf{Architectures.} We incorporate a convolutional network for all modules except the transformer-based ones and the trajectory encoder. The discriminator architecture is multi-scale and multi-patch, inspired by prior work~\citep{isola2017image}. The dimension of the latent vector is set at $z \in \R^{64 \times 4 \times 4}$ which is split in 4 patches resulting in the flattened embedding size of 256. For the prediction network, the decoder and variational modules each consist of 6 layers and 6 heads, collectively comprising 16.1 million parameters. The image module employs a DINOv2 ViT-S/14 backbone~\citep{oquab2023dinov2}. Within this module, the image, spatial, and temporal layers comprise 1, 4, and 4 layers, respectively, each with 4 heads. The total number of parameters for the image head is 27.4M. For the diffusion-based decoder, we leverage an implementation by HuggingFace~\cite{von-platen-etal-2022-diffusers}. In the results section, we use a single-step decoder, unless stated otherwise. 

\textbf{Model Training.} We used the AdamW optimizer \citep{loshchilov2017decoupled} with $lr=\num{4e-4}$.
For representation learning, we used three NVIDIA TITAN X 24 GB for 360k steps with a total batch size of 30. We trained the L-OGM-only prediction models on a single NVIDIA TITAN RTX 24GB GPU and the multimodal model and batch decoder on two NVIDIA L40 48GB GPUs. The models were trained with a total batch size of 40 until convergence.
For the stochastic prediction component, we use a KL annealing with $\beta=\num{2e-6}$ for the first 10 epochs followed by a linear increase to 0.2 over 50k training steps.

\subsection{Evaluation}
\label{sec:evaluation}
\textbf{Baselines.} We benchmark our approach against methods commonly used in L-OGM prediction, including PredNet~\citep{lotter2016deep, itkina2019dynamic} and TAAConvLSTM~\citep{lange2020attention}. Additionally, we compare against OccWorld~\citep{zheng2023occworld}, a state-of-the-art network for 3D semantic occupancy prediction adapted for the 2D L-OGM task. OccWorld uses the latent space of VQ-VAE and incorporates a customized transformer to enhance inference efficiency. To ensure a fair comparison, we increase its latent space size to match the reconstruction performance of our method. We further evaluate our approach against state-of-the-art real-time video prediction models, including SimVP V2~\citep{tan2022simvp}, PredRNN V2~\citep{wang2022predrnn}, and E3DLSTM~\citep{wang2019eidetic}. We train all models until convergence. As an additional baseline, we include a naive method that repeats the last observed frame across the entire prediction horizon, providing a reference for assessing the models’ ability to capture the scene’s motion dynamics.

\textbf{Metric}: We evaluate all models using the Image Similarity (IS) metric~\citep{birk2006merging} across the \SI{1.5}{\second} and \SI{3.0}{\second} prediction horizons. 
For stochastic predictions, we sample 10 predictions and evaluate the best one. IS calculates the smallest Manhattan distance between two grid cells with the same thresholded occupancy, capturing the spatial error of predictions. It determines the grid distance between matrices $m_1$ and $m_2$~\cite{birk2006merging}: 
\begin{equation}
\begin{split}
    \psi(m_1,m_2) = \sum_{c \in \mathcal{C}} d(m_1,m_2,c) + d(m_2,m_1,c) 
\end{split}
\end{equation}
where
\begin{equation}
\begin{split}
    d(m_1,m_2,c) = \frac{\sum_{m_1[p]=c} \text{min} \{\text{md}(p_1,p_2)|m_2[p_2]=c\}}{\#_c(m_1)}.
\end{split}
\end{equation}
The set of discrete values $\mathcal{C}$ possibly assumed by $m_1$ or $m_2$ are: occupied, occluded, and free. $m_1[p]$ denotes the value $c$ of map $m_1$ at position $p=(x,y)$. $\text{md}(p_1,p_2)=|x_1 - x_2| + |y_1 - y_2|$.  $\#_c(m_1)= \#\{p_1 \mid m_1[p_1]=c\}$ is the number of cells in $m_1$ with value $c$.

\section{Results}
\label{sec:results}
\subsection{Latent Space Analysis}
\label{sec:latent_analysis}
\input{Figures/recon_v2}
The latent space trained during the representation learning stage is crucial for facilitating accurate predictions. 
If an agent in the observed frames is lost during the encoding phase, the prediction network struggles to recover this information, leading to incorrect forecasts. 
We examine the impact of different representation learning losses on reconstruction and prediction performance in \cref{tab:latent_space_table}. Our results highlight a trade-off between single-grid reconstruction performance and prediction performance. The simple autoencoder performs well in reconstruction and short-term prediction but suffers in prediction accuracy as the horizon extends. In contrast, incorporating KL and adversarial components reduces reconstruction performance but enhances long-term prediction capability, aligning with the primary objective of this work. We also report the performance of our representation learning approach with all augmentations, demonstrating its positive impact on all metrics. For context, we provide randomly selected L-OGMs and their reconstructions in \cref{fig:recon}, which illustrate that the encoder-decoder setup effectively reconstructs the scenes.

\begin{figure*}[t]
  \centering
  \scalebox{0.985}{
  \begin{tikzpicture}
    \node (img1) at (-6.2,0) {\input{Figures/prediction-0082-270}};
    \node (img2) at (0.1\textwidth,0) {\input{Figures/new_prediction-0551-30}};
  \end{tikzpicture}
  }
    \caption{Examples of LOPR and OccWorld-2D predictions with visualized front camera observations from the nuScenes dataset. The predictions are generated with the single-step decoder. LOPR is conditioned on all cameras around the vehicle, maps, and the planned trajectory. We report IS scores for each sample. (Left) Prediction of an oncoming vehicle (red) visible only in the front camera. Each LOPR sample captures a realistic hypothetical evolution of the scene, such as variations in the velocity of the oncoming car. (Right) Correct forecasting of an oncoming vehicle and a static road layout (orange).
    Both examples demonstrate that our framework is capable of multi-future reasoning and leveraging multi-modal observations.}
\vspace{-0.4cm}
\label{fig:pred1}
\end{figure*}
\begin{figure*}[ht]
  \centering
    \scalebox{0.985}{
  \begin{tikzpicture}
    \node (img1) at (-6.2,0) {\input{Figures/new_prediction-0266-310}};
    \node (img2) at (0.1\textwidth,0) {\input{Figures/prediction-0266_30}};
  \end{tikzpicture}
  }
  \caption{Examples of LOPR and OccWorld-2D predictions with visualized front camera from the nuScenes dataset. The prediction setting is the same as in \cref{fig:pred1}. (Left) Accurate prediction of the static road layout and a parked vehicle (green) visible only in the front camera. Each LOPR sample provides a realistic occupancy representation of the parked car and the environment's layout. (Right) Accurate prediction of a parked truck (blue).}
\label{fig:pred2}
\end{figure*}
\begin{table}[t]
\setlength{\tabcolsep}{2.5pt} 
 \centering
 \caption{The impact of representation learning objective on the reconstruction ($\textbf{IS}_{recon}$) and prediction performance ($\textbf{IS}_{5\rightarrow X}$) on nuScenes with no augmentation during representation learning. VAE-GAN (KL + Adv) excels at long-term prediction horizon. We also report VAE-GAN performance with the augmented dataset.}
    \scalebox{0.90}{
    \begin{tabular}{cccccc}
        \toprule
        Recon & KL & Adv & $\textbf{IS}_{recon}(\downarrow)$ & $\textbf{IS}_{5\rightarrow15}(\downarrow)$ & $\textbf{IS}_{5\rightarrow30}(\downarrow)$ \\
        \midrule
        \multicolumn{5}{l}{No augmentations in representation learning} \\
        \midrule 
        $\checkmark$ & $\times$ & $\times$ & \textbf{2.18 $\pm$ 0.01} & \textbf{7.88 $\pm$ 0.16} & 12.18 $\pm$ 0.29 \\ 
        $\checkmark$ & $\checkmark$ & $\times$ & 5.90 $\pm$ 0.01 & 9.76 $\pm$ 0.15 & 12.69 $\pm$ 0.20 \\ 
        $\checkmark$ & $\times$ & $\checkmark$ & 4.36 $\pm$ 0.01 & 8.36 $\pm$ 0.14 & 11.72 $\pm$ 0.21 \\ 
        $\checkmark$ & $\checkmark$ & $\checkmark$ & 4.82 $\pm$ 0.01 & 7.94 $\pm$ 0.12 & \textbf{10.56 $\pm$ 0.17}\\
        \midrule
        \multicolumn{5}{l}{With augmentations in representation learning} \\
        \midrule
        $\checkmark$ & $\checkmark$ & $\checkmark$ & 3.30 $\pm$ 0.01 & 7.00 $\pm$ 0.10 & 9.76 $\pm$ 0.16\\
        \bottomrule
    \end{tabular}}
    
    \label{tab:latent_space_table}
\end{table}

\begin{table}[t]
    \setlength{\tabcolsep}{1pt} 
    \centering
    \caption{Quantitative evaluation of the prediction performance. The best results are bolded, and the second-best results are underlined. $\dagger$ indicates a stochastic model. In \textit{Ours}, $+$ indicates the addition of a feature to a model that includes all modifications listed in the rows above. We also report the frequency and CUDA memory consumption for generating a single 15-frame prediction.}
    \tiny 
    \scalebox{1.35}{
    \begin{tabular}{@{}lcccccccc@{}}
      \toprule
      \textbf{Model} & \multicolumn{2}{c}{\textbf{NuScenes Dataset}} & \multicolumn{2}{c}{\textbf{Waymo Open Dataset}} & \\ 
      & $\textbf{IS}_{5 \rightarrow 15}$ & $\textbf{IS}_{5 \rightarrow 30}$  &
      $\textbf{IS}_{5 \rightarrow 15}$  & $\textbf{IS}_{5 \rightarrow 30}$ & T (Hz) & M (GB)  \\
      \midrule
      PredRNN V2 & 30.69$\pm$2.10 & 79.95$\pm$4.07 & 28.45$\pm$1.98 & 62.71$\pm$3.38 & 7.69 & 0.23 \\
      Sim.VP V2 & 20.02$\pm$1.28 & 47.20$\pm$2.63 & 15.87$\pm$1.18 & 46.38$\pm$2.66 & \textbf{37.04} & 0.26  \\
      ED3LSTM & 10.33$\pm$0.29 & 19.49$\pm$0.71 & 14.36$\pm$1.04 & 36.60$\pm$2.15 & 5.55 & 0.94  \\
      TAAConvLSTM & 7.02$\pm$0.17 & 15.26$\pm$0.66 & 6.43$\pm$0.36  & 20.51$\pm$1.39 & 5.65 & 0.11  \\
      PredNet & 7.10$\pm$0.19 & 13.93$\pm$0.44 & 6.78$\pm$0.41 & 22.38$\pm$1.54 & 16.39 & \textbf{0.10}  \\
      OccWorld-2D$^{\dagger}$ & 7.84$\pm$0.09 & 11.90$\pm$0.18 & 
      6.72$\pm$0.13 & 11.03$\pm$0.26 & 3.00 & 1.10 \\     
      Fixed Frame & 11.50$\pm$0.14 & 14.41$\pm$0.18 &10.35$\pm$0.41 & 14.74$\pm$0.54 & - & -   \\
      \midrule
      Ours \\
      \midrule
      Deterministic  & 7.94$\pm$0.13 & 11.48$\pm$0.22 & 7.62$\pm$0.33 & 12.12$\pm$0.73 & \underline{25.64} & 2.49 \\
      $+$Aug.  & 7.24$\pm$0.11 & 10.47$\pm$0.19 & 6.65$\pm$0.26 & 10.60$\pm$0.53 & \underline{25.64} & 2.49 \\
      $+$Stoch.$^{\dagger}$  & 7.00$\pm$0.10 & 9.76$\pm$0.16 & 6.64$\pm$0.19 & 9.93$\pm$0.28 & 11.91 & 2.51 \\
      $+$R+T+M$^{\dagger}$  & \underline{6.36}$\pm$\underline{0.08} &  \underline{8.32}$\pm$\underline{0.12}  & \underline{6.23}$\pm$\underline{0.18} & \underline{9.00}$\pm$\underline{0.28} & 9.09 & 2.61  \\
      $+$Diff. Dec.$^{\dagger}$   & \textbf{4.88}$\pm$\textbf{0.09} & \textbf{7.12}$\pm$\textbf{0.12} & \textbf{5.24}$\pm$\textbf{0.18} & \textbf{8.17}$\pm$\textbf{0.27} & $<$0.01 & 17.99 \\
      \bottomrule
    \end{tabular}}
  \label{tab:results}
\vspace{-0.3cm}
\end{table}

\subsection{Prediction Task}
\label{sec:pred_task}
\textbf{General Performance}: We compare our framework against the baselines in \cref{tab:results}. The integration of dataset augmentations, stochasticity, additional input modalities, and a diffusion decoder each contribute to notable improvements. LOPR outperforms previous methods on both datasets, with improvements becoming more pronounced as the prediction horizon extends. We visualize examples of predictions rolled out for \SI{3.0}{s}, extending beyond the prediction horizon used during training, in \cref{fig:pred1,fig:pred2}. Our framework generates high-quality, realistic predictions, supporting the quantitative results while remaining real-time feasible. 

\textbf{Stochasticity}: Stochasticity in the L-OGM primarily stems from the unknown intents of other agents in the scene and partial observations. Our framework models the multimodal distribution of future agents' positions. It is capable of inferring hypothetical, previously unobserved agents (see \cref{fig:stochastic}) and capturing the varying motion dynamics of observed agents (see \cref{fig:pred1,fig:pred2,fig:stochastic}). As shown in \cref{tab:results}, modeling stochasticity has a positive quantitative impact compared to the deterministic framework. The variational module enables high-quality and semantically meaningful multi-future reasoning about the behaviors of both observed and unobserved agents.



\textbf{Impact of trajectory, map, and camera conditioning}: 
LOPR leverages additional input modalities to enhance its understanding of the surroundings and the ego vehicle’s intent, enabling more accurate predictions. This includes accurately inferring static elements of the environment, such as road layouts and parked vehicles, as well as dynamic components like oncoming vehicles. These features, while visible in the cameras and maps, lie beyond the observable area in L-OGMs (see \cref{fig:pred1,fig:pred2}).
In \cref{tab:modality_ablation}, we evaluate the impact of incorporating trajectory conditioning, maps, and cameras as input modalities. Our results demonstrate that each modality contributes to significant numerical improvements, with the best performance achieved when all available input modalities are combined. Furthermore, we observe a positive impact of DINO-based camera module finetuning on the deterministic occupancy prediction task. 
\begin{table}[h]
    \setlength{\tabcolsep}{1pt} 
    \centering
    \caption{Impact of trajectory, map, and camera modalities on the prediction performance. $*$ denotes a camera module without finetuning on the deterministic task. All models were trained for 40 epochs.}
    \tiny 
    \scalebox{1.6}{
    \begin{tabular}{ccccc}
        \toprule
        Traj. & Maps & Cameras & $\textbf{IS}_{5\rightarrow15}(\downarrow)$ & $\textbf{IS}_{5\rightarrow30}(\downarrow)$ \\
        \midrule
        $\times$ & $\times$ & $\times$ & 7.20 $\pm$ 0.11 & 10.02 $\pm$ 0.16 \\ 
        $\checkmark$ & $\times$ & $\times$ & 6.75 $\pm$ 0.11 & 8.94 $\pm$ 0.14 \\ 
        $\times$ & $\checkmark$ & $\times$ & 6.88 $\pm$ 0.10 & 9.18 $\pm$ 0.14 \\
        $\times$ & $\times$ & $\checkmark$ & 6.91 $\pm$ 0.10 & 9.54 $\pm$ 0.15  \\
        $\times$ & $\times$ & $\checkmark^{*}$ & 7.20 $\pm$ 0.11 & 9.98 $\pm$ 0.16 \\ 
        $\checkmark$ & $\checkmark$ & $\times$ & 6.73 $\pm$ 0.10 & 8.86 $\pm$ 0.14 \\
        $\checkmark$ & $\times$ & $\checkmark$ & 6.47 $\pm$ 0.09 & 8.61 $\pm$ 0.13 \\ 
        $\times$ & $\checkmark$ & $\checkmark$ & 6.62 $\pm$ 0.09 & 8.83 $\pm$ 0.13 \\ 
        $\checkmark$ & $\checkmark$ & $\checkmark$ & \textbf{6.44} $\pm$ \textbf{0.09} & \textbf{8.50} $\pm$ \textbf{0.12} \\
        \bottomrule
    \end{tabular}}  
  \label{tab:modality_ablation}
  \vspace{-0.5cm}
\end{table}

\textbf{Diffusion-based Decoder}: Known weaknesses of making predictions in the latent space of generative models include poor temporal consistency and compression losses. 
We address these concerns with a diffusion-based batch decoder which leads to significant improvements in temporal consistency and the maintenance of detail that is often lost due to compression, as shown in \cref{fig:diffusion}. Numerically, it also results in significant improvements, especially in short-horizon predictions, due to the preservation of static details. However, this comes at the cost of real-time feasibility. While our approach with the single-step decoder and all baselines generate a full sequence of predictions at a rate of \SI{3}-\SI{38}{Hz}, decoding predictions with a diffusion decoder takes minutes. Although it is not currently feasible for real-time applications, recent advancements in diffusion-based consistency models~\citep{song2023consistency} suggest a promising path toward achieving this.
\input{Figures/scene-0343_70}
\input{Figures/diffusion}

\textbf{Real-time Feasibility}: In \cref{tab:results}, we provide the frequency of generating a single 15-frame prediction in full precision and the CUDA memory consumption for context. All models are tested on NVIDIA L40. Our results demonstrate that LOPR with a single-step decoder is real-time feasible, achieving \SI{11.91}{Hz} without additional input modalities and \SI{9.09}{Hz} when all input modalities are used. In comparison, OccWorld-2D, which operates in the latent space of VQ-VAE and uses a customized transformer to accelerate inference, achieves \SI{3.00}{Hz}. For further context, vanilla transformer applied to a similar size discrete latent space and observation-prediction horizon requires 30 seconds to generate a single prediction on an NVIDIA V100~\citep{rakhimov2020latent}.


\textbf{Baseline Comparison}: There are several reasons for the significant improvements over prior work. Most prior work is optimized for deterministic prediction tasks in grid cell space and fails to condition on the input modalities available in the autonomous vehicle stack. These methods do not capture the stochastic nature of motion forecasting and lack semantic information about the scene (e.g., distinguishing whether an occupied cell is due to a pedestrian or a sign). As a result, their forecasted L-OGMs gradually lose important details over the prediction horizon, leading to the disappearance of moving objects. While they might capture some static details of the scene, resulting in respectable IS scores for short prediction horizons, they ultimately yield high IS scores as static details vanish and deterministic predictions prove insufficient. Our numerical results demonstrate that the deterministic variation of our framework, conditioned on L-OGMs (the same setting as the baselines), outperforms all deterministic baselines over extended prediction horizons. As we incorporate multi-future reasoning, provide additional semantic conditioning, and introduce diffusion decoding, the performance gap between LOPR and the baselines becomes even more substantial, as shown in \cref{tab:results}. Interestingly, our work also outperforms the state-of-the-art in semantic occupancy prediction, which uses a customized transformer to predict discrete latent representations. This highlights an alternative to the commonly used transformer with discrete codebooks, which often incurs significant cost at inference time.
\section{Limitations}
\label{sec:limitations}
LOPR relies on a well-trained latent space, which may sometimes lose critical details necessary for accurate predictions. If the latent encoding misses important information, the prediction network may not recover it, leading to potentially inaccurate predictions. We have partially addressed this issue with a diffusion decoder. However, this approach may still result in inaccuracies and is not real-time feasible. 
Additionally, the performance of our framework is heavily influenced by the size of the available perception datasets. The datasets used in this paper contain about six hours of data, which limits performance in scenarios not well-represented in the dataset, such as intersection interactions and cross-traffic.

\section{Conclusion}
\label{sec:conclusion}
In this paper, we proposed a self-supervised L-OGM prediction framework that captures the stochasticity of the scene and is conditioned on multi-modal observations available in autonomous vehicles. LOPR consists of a VAE-GAN-based generative model that learns an expressive low-dimensional latent space, and a transformer-based stochastic prediction network that operates on this continuous latent space.
Our experiments demonstrate that LOPR outperforms all prior approaches both qualitatively and quantitatively while maintaining real-time feasibility. Furthermore, it surpasses commonly used transformer-based methods with discrete representations, while offering significantly faster inference. We also highlight the advantages of incorporating trajectory, map, and camera conditioning, which enhance the framework’s capabilities. Additionally, we extend the framework with a diffusion decoder to address temporal consistency issues and reduce compression losses, albeit with a trade-off in real-time feasibility.
In future work, we will explore extending LOPR to perform 3D occupancy prediction, and apply it to other tasks, such as occlusion inference \cite{itkinaicra2022, lange2023scene} and path planning.

\section*{Acknowledgments}
This project was made possible by funding from the Ford-Stanford Alliance.

\bibliographystyle{plainnat}
\bibliography{references}

\end{document}